# Метод автоматического построения онтологии предметной области на основе анализа лингвистических характеристик текстового корпуса

Е.А. Оробинская

Лионский университет Лион-2 им. Люмьер, Франция
Olena.Orobinska@univ-lyon2.fr

**Аннотация**

Вызов сегодняшнего дня, обусловленный бурным ростом текстовых хранилищ в глобальной сети Интернет, заключается в том, чтобы «научить» сами информационные системы (ИС) обнаруживать и правильно интерпретировать полезную информацию, предоставляемую текстом.

Разработка методов автоматического анализа текстовой информации и извлечение из полнотекстовых документов релевантных данных является актуальной задачей инженерии знаний в целом и онтологического инжиниринга в частности. Предпосылкой появления прорывных технологий, которые позволи бы ее решить автор считает поиск решений на основе методов системного анализа, как самого объекта исследования – текста, так и поставленных прикладных задач, которые должны быть решены в результате такой обработки. Иными словами, для семантического анализа текстовой информации требуется онтологический подход, т.е. чтобы обнаружить в тексте требуемую информацию, сама ИС должна обладать достаточным объемом лингвистических знаний.

## Введение

Со ссылкой на [2],[3] можно утверждать, что в 2011 г. наиболее быстро растущей категорией т. наз. "sweet tools" или интеллектуальных систем были именно онтологии. Обобщая множество сходных определений, напомним, что в настоящее время под онтологией подразумевается семантическая структура, в явном виде описывающая основные понятия (концепты) предметной области (ПО), связывающие их отношения, а также аксиомы, описывающие правила вывода новых концептов [4].

До сих пор, построение онтологии остается приоритетной задачей экспертов ПО, что делает этот процесс очень дорогостящим и долгосрочным. Именно этими факторами объясняется большое количество работ, появившихся в последнее время и предлагающих новые методы анализа текста и, на их основе, технологии автоматизированного построения онтологий [2].

До начала 2000–ных годов для решения задач анализа текста в основном применялись и развивались статистические методы. С появлением сетевых лексических ресурсов и совершенствованием синтаксических анализаторов линвистические методы привлекают все большее внимание специалистов благодаря высокому качеству результатов, которые обещают такие методы [5]. Поэтому целью данной работы является описание структуры ИС, позволяющей обнаруживать в текстовом корпусе релевантные концепты ПО (радиационная защита) на основе использования лингвистических шаблонов, характерных для данной ПО.

## Основные принципы построения онтологий

Автор в основном разделяет подход к построению онтологии на основе текста, предложенный Buitelaar и коллегами в 2005 г. [6], известный как Layer Cake. Весь процесс построения онтологии разбит на несколько независимых этапов, на каждом из которых решается одна определенная задача, результаты которой, в свою очередь, служат исходными данными для задачи следующего, как правило, более сложного уровня. Согласно этому подходу можно выделить следующую последовательность действий: извлечение из текста терминов-кандидатов → разбиение терминов на группы (кластеризация) → присвоение обобщающего понятия-концепта каждой группе → определение отношений между концептами → формирование правил вывода (расширения концептов). Для начальных этапов многочисленные статистические методы позволяют получать весьма качественные результаты (до 90% точности по сравнению с результатами работы экспертов [7]).

Однако для построения таксономии понятий и обнаружения неткасономических отношений необходима разработка лингвистических методов. В лингвистических шаблонов, предложенных еще М. Hearst [8]. Дальнейшее совершенствование такого подхода зависит от наличия общей теории описания и анализа конкретного языка. В этой связи автор разделяет идею Золотовой Г.А. о том, что синтаксический строй речи (текста) организуется регулярными комбинациями «элементарных» единиц, из которых строятся все другие более сложные (речевые или текстовые) конструкции [9]. В качестве такой единицы выдвинуто понятие синтаксемы. Синтаксемой по Золотовой Г.А. называется минимальная далее неделимая семантико-синтаксическая единица русского языка, выступающая одновременно и как носитель элементарного смысла, и как конструктивный компонент с функциональностью, необходимой и достаточной для построения более сложных синтаксических конструкций.

Признаками синтаксемы служат: 1) категориальное значение слова, от которого она образована, 2) определенная (конкретная) морфологическая форма, 3) функциональность, вытекающая из двух первых признаков, как способность реализовываться в определенных позициях (речи, теста), т.е. как возможная роль в построении коммуникативной единицы. Золотова Г.А. различает три таких функциональных роли:

1) возможность самостоятельного независимого существования синтаксемы;

2) использование синтаксемы в качестве компонента предложения;

3) использование синтаксемы в качестве компонента словосочетания (или сочетания слов), так называемое присловное употребление.

Функциональные свойства синтаксем служат основой для разграничения трех их возможных типов, называемых соответственно, свободными – А (или обладающими полным функциональным репертуаром), обусловленными – Б (способными выполнять функции 2, 3, крайне изредка –1), связанными – В (выступающими только в роли 3). Очевидно, что функциональные свойства синтаксем имеют транзитивный характер.

Рассматривая текст как динамическую систему, мы имеем возможность описать весь процесс построения онтологии как целенаправленную операционную деятельность в пределах данной системы, организованную для решения задач содержательного наполнения элементов онтологии. Математически онтология определяется следующим образом [6]:

$$O := (C, \leq_c, R, \sigma_R, \leq_R, A, \sigma_A, T),$$

где:

частности, одним из эффективных подходов для обнаружения группы синонимов и общего для них гиперонима является использование

1. C, R, A, T – являются несвязанными множествами, чьи элементы называются идентификаторами концептов, отношений, атрибутов и типов данных соответственно;
2. $\leq_c$ – полусвязанная таксономия (semi-upper lattice) концептов с общим элементом самого верхнего уровня rootC;
3. функция σR:R→C+ называемая признаком отношения (relation signature);
4. $\leq_R$ on R, иерархия отношений где, r1 $\leq_R$ r2 подразумевает =|σR(r1)| =|σR(r2)| и πi(σR(r1)) $\leq_c$ πi(σR(r2)) для каждого $\leq$ i $\leq$ =|σR(r1)|;
5. функция σA:A→C×T называемая признаком атрибута (attribute signature);
6. множество типов данных (datatypes) T, таких как строки, целые числа и т.д.

Для обеспечения языковой компетентности, достаточной для построения и самообучения онтологии предметной области на базе текста, ИС сама должна обладать знаниями соответствующего порядка, общими (языковыми) и специальными, (относящимися к данной ПО). Такая ИС должна, по сути, объединять в себе две «встроенные» онтологии: онтологию языка (русского) и базовую (стартовую) онтологию ПО. Многочисленные электронные словари-тезаурусы (MRD – machine readable dictionary), с лексикографической разметкой (какие как WordNet и его русскоязычные аналоги RussNet, RuNet и др.) можно рассматривать как элементы C и $\leq_c$ требуемой лингвистической онтологии. Парсеры языка (например, проект АОТ [6]) могут быть успешно использованы как для определения ролей концептов, так и для определения их атрибутов. То, чего действительно не хватает для полноценной интеллектуальной системы, так это систематизированного репертуара лексико-синтаксических единиц языка, несущих в себе однозначно трактуемую семантику и одновременно выполняющих роль «элементарных единиц сборки» высказываний (текстов). Пересечение такого репертуара «архе-функций», несомых синтаксической формой и репертуара слов-носителей категориальных референций, дает проекцию однозначно трактуемой роли (функции) выполняемой данным понятием. Подобный систематизирующий труд, как было сказано выше, выполнен в значительной степени Золотовой Г.А. [8] и ждет достойной технической реализации.

## ИС для автоматизированного обнаружения концептов онтологии ПО

Предлагаемый подход автоматизированного построения онтологии ПО, сочетает в себе быстроту статистических методов и точность лингвистического подхода с позиций синтаксем (или лингвистических шаблонов).

Процесс построения онтолгии состоит из нескольких последовательных этапов:
1. Разметка корпуса синтаксическими тегами (напрмер, информацией о части речи, роде числе и т.д.).
2. Определение характерных, т.е. наиболее часто встречающихся в корпусе лингвистических структур и формирование на их основе лигвистических шаблонов.
3. Поиск фрагментов текста, соответствующих данным шаблонам.
4. Нормализация и добавление в базовую онтологию новых концептов, выявленных на предыдущем этапе.

На третьем этапе необходимо привлечение эксперта предметной области для оценки релевантности найденных потенциальных концептов.

Проиллюстрируем сказанное примером анализа следующего текста (фрагмент рекомендации МАГАТЭ 2000 г.):

«После аварийного облучения большой мощности развитие поражений по времени проходит через четко определенные стадии. Продолжительность и время наступления стадий зависит от дозы. Малые дозы не дают наблюдаемых эффектов. Типичное развитие событий после облучения всего тела от источника проникающей радиации включает начальную продромальную стадию с такими симптомами, как тошнота, рвота, усталость и, возможно, лихорадка и диарея, после чего следует латентный период различной продолжительности.»

Модуль лингвистической онтологии ИС распознает такие свойства концептов: квалитатив (качество) субъекта и его объект. В русском языке это может выражаться следующими формулами:
1. вспомогательными глаголами {отличаться, выражаться, проявляться, характеризоваться, обладать} + Т.п. признакового слова (предицирующий компонент);
2. с помощью родительного падежа, когда предицирущий компонент определен сочетанием {прилагательное + существительное};
3. в эксплицитной форме конструкцией типа «объект может иметь такие {характеристики, свойства, признаки}».
4. с помощью связанной синтаксемы с родительным падежом, обозначающей объект при отглагольном существительном.

Обнаружение синтаксических характеристик выполняется прасером русского языка (АОТ-парсер [10]), способным распознавать грамматические формы встречаемых слов. Для правильной работы определяется также порядок применения (приоритет) правил.

Определяются свойства концепта «облучение», т.е. система применяет описанные правила только к тем предложениям, в которых обнаружит любую словоформу «облучение».

Таким образом, имеем возможность обнаружить в тексте и истолковать, например, следующие свойства термина:

«…облучения большой мощности» – правило 2; «мощность» – атрибут концепта «облучение»;

«…облучение всего тела» – правило 4; предикат «облучать» и его объект «тело»;

Преимуществом предлагаемого метода является то, что «точками входа» для анализа текста могут служить и слова и собственно синтаксемы.

Структура ИС представлена на рис.1:

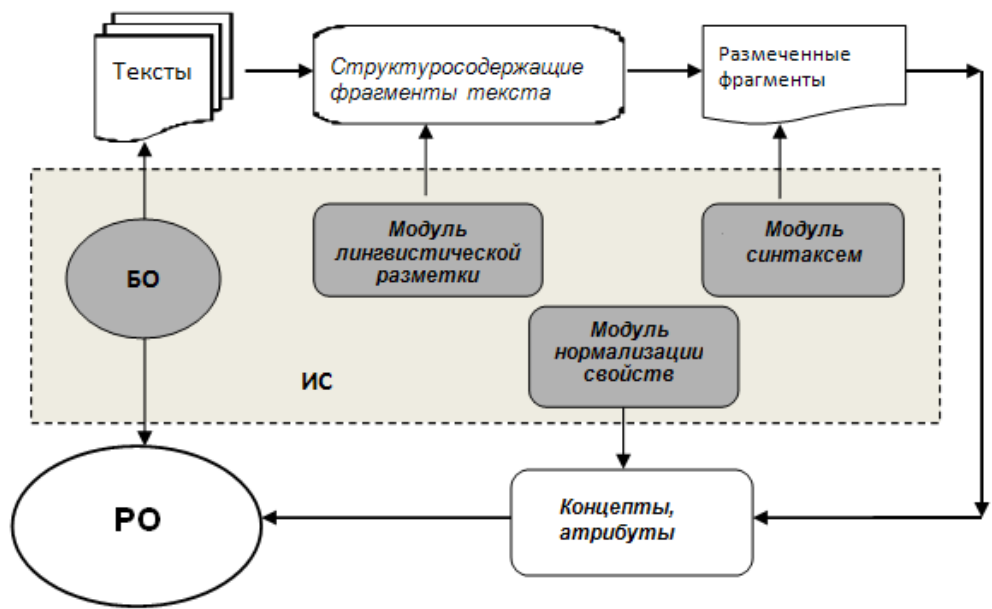

**Рис.1.** Фреймворк ИС на основе синтаксем

ИС состоит из следующих модулей:

Модуль базовой онтологии (БО) построен вручную с помощью эксперта ПО.В настоящий момент базовая онтология, представленная на OWL содержит около 30 концептов и описывающих их отношений. БО одновременно служит «фильтром» для отбора фрагментов текста, потенциально содержащих информацию, которая позволит либо добавить новые свойства для существующих концептов БО, либо добавить новые концепты в БО (с их свойствами) и получить расширенную онтологию (РО).

Модуль лингвистической разметки, основой которого служит парсер языка. На выходе этого модуля получаем фрагменты текста, снабженные необходимой синтаксической информацией для дальнейшей обработки.

Модуль синтаксем русского языка (основной модуль ИС), включающий в себя словарь типа WordNet (аналоги для русского языка RuNet, RussNet), объединяющий слова в группы синонимов.

Модуль нормализации свойств, преобразующий найденные на предыдущем этапе данные в формальное представление на OWL и обеспечивающий расширение БО.

## Заключение

Анализ текстовой информации и извлечение из полнотекстовых документов релевантных данных остается актуальной задачей онтологического инжиниринга в частности. Качественное расширение возможностей ИС возможно при условии внедрения в них модулей, способных извлекать характеристики концептов на основе лингвистического анализа. Одним из возможных способов решения этой задачи является использование лингвистических шаблонов. В данной работе предложен подход к автоматизированному построению (расширению) базовой онтологии на основе синтаксем русского языка. Поскольку каждая синтаксема описывается конечным детерминированным множеством признаков, такой подход является не только возможным, но и предпочтительным, поскольку он обеспечивает однозначное определение свойств концептов создаваемой онтологии. Трудоемкость задачи окупается качеством получаемых результатов. В будущем планируется проверка эффективности метода на большом корпусе, а также возможность его применения для анализа текстов на других языках (аглийский, французский).

## Литература

# Automatic Method Of Domain Ontology Construction based on Characteristics of Corpora POS-Analysis

O. Oroinska

It is now widely recognized that ontologies, are one of the fundamental cornerstones of knowledge-based systems. What is lacking, however, is a currently accepted strategy of how to build ontology; what kinds of the resources and techniques are indispensables to optimize the expenses and the time on the one hand and the amplitude, the completeness, the robustness of en ontology on the other hand. The paper offers a semi-automatic ontology construction method from text corpora in the domain of radiological protection. This method is composed from next steps: 1) text annotation with part-of-speech tags; 2) revelation of the significant linguistic structures and forming the templates; 3) search of text fragments corresponding to these templates; 4) basic ontology instantiation process.